# Novel Deep Learning Model for Traffic Sign Detection Using Capsule Networks

Amara Dinesh Kumar
Department of Electronics and Communication Engineering
Amrita School of Engineering, Coimbatore
Amrita Vishwa Vidyapeetham, India
cb.en.p2ael17001@cb.students.amrita.edu

R. Karthika
Department of Electronics and Communication Engineering
Amrita School of Engineering, Coimbatore
Amrita Vishwa Vidyapeetham, India
r_karthika@cb.amrita.edu

Latha Parameswaran
Department of Computer Science and Engineering
Amrita School of Engineering, Coimbatore
Amrita Vishwa Vidyapeetham, India
p_latha@cb.amrita.edu

*Abstract*—*convolutional neural networks are the most widely used deep learning algorithms for traffic signal classification till date[1] but they fail to capture pose, view, orientation of the images because of the intrinsic inability of max pooling layer. This paper proposes a novel method for Traffic sign detection using deep learning architecture called capsule networks that achieves outstanding performance on the German traffic sign dataset. Capsule network consists of capsules which are a group of neurons representing the instantiating parameters of an object like the pose and orientation[2] by using the dynamic routing and route by agreement algorithms. unlike the previous approaches of manual feature extraction, multiple deep neural networks with many parameters, our method eliminates the manual effort and provides resistance to the spatial variances. CNNs´ can be fooled easily using various adversary attacks[3] and capsule networks can overcome such attacks from the intruders and can offer more reliability in traffic sign detection for autonomous vehicles. Capsule network have achieved the state-of-the-art accuracy of 97.6% on German Traffic Sign Recognition Benchmark dataset (GTSRB).*

*Keywords: CNN, Capsule Net, Pose, Traffic sign, Dataset*

## I. INTRODUCTION

Traffic sign detection is a real world task which involves lot of constraints and complications. Even a minor misclassification of the traffic sign can lead to catastrophic consequences and can even lead to loss of life. It is implemented in various advanced driver assistance systems and in autonomous vehicles. A camera is present on the dashboard of the vehicle and it captures the real time video feed which is sampled into frames and fed to a deep learning model which is deployed inside a automotive embedded board. As the vehicle is driven in various environments, lighting conditions, speeds and geographies it is essential for the deep learning algorithm to be robust and reliable at all times. The camera can capture the traffic sign in different orientations and poses but the algorithm should be able to recognize the correct sign[4] and capsule networks are the perfect deep learning algorithm in addressing this problem.

Generally Convolutional neural networks are used for all the state of the art deep learning neural network algorithms[5] in most of the image related tasks. Convolution captures the spatial information of the image using the kernel function in convolution layer. A CNN consists of input, output and hidden layers. The hidden layers further consists of convolutional, pooling, fully connected and normalization layers. CNNs perform very well for image related operations but they have some fundamental limits and draw backs. The CNN fail[6] to capture the relative spatial and orientation relationships. CNN can get confused easily by image orientation or by change in pose.

Pose information can be rotation, thickness, skew, precise object position. Max polling is the biggest drawback for the CNN as they cannot propagate the spatial hierarchies between simple and complex objects which lead to invariance and makes them fail to capture the pose and the spatial relationships between the pixels in the image. CNN uses the max pooling layer which down samples the data and reduces the spatial information of data that is passed to the next layer. To overcome this drawback capsnet architecture is invented which reached the state of the art performance on MNIST digits dataset[2] and obtained better results than CNNs on Multi MNIST dataset.

## II. RELATED WORK

Comparing the research work done in the past is difficult in the traffic sign detection area because of the extensive research effort carried by researchers in this area and use of different types of datasets for solving different set of problems which includes the detection, classification and tracking related tasks.

### A. Using computer vision feature extraction methods

This is one of the early approach under which a lot of algorithms and methodologies were proposed by the computer vision scientists before the advent of machine learning. Techniques like Histogram Oriented Gradients ( HOG )[7] is initially popularized for detection of pedestrians. In this method gradients of color images are computed along with different normalized, weighted histograms. Scale Invariant Feature Transform (SIFT)[8] technique was used for classification and sliding window approach was used to perform both classification and detection tasks simultaneously.

### B. Using machine learning

Several kinds of machine learning algorithms[9] like support vector machines, linear discriminant analysis[10], ensemble classifiers, random forest and kd-trees[11] were used for the traffic sign classification.

Linear Discriminate Analysis(LDA)[5] is based on maximum posteriori estimation of the class membership. Class densities are assumed to have multi variate Gaussian and common co-variance matrix.

Random Forest is a ensemble classifier method[1] which is based on the set of non pruned random decision trees. Each decision tree is built using the randomly took training data.For classification testing data is evaluated by all the decision trees and the classification output is based on the majority voting considering decisions of all the majority decision trees.

Support Vector Machines(SVM) is a classification algorithm which classifies the data by dividing the n dimensional data plane with a hyper plane for classification[9].SVM can even separate non-linearly scattered data by transforming the classification plane to higher dimensions using non linear kernel function which uses a method called kernel trick for its implementation.

Machine learning approaches[12] were unable to handle different aspect ratio and variable size images and features have to be manually hand engineered which is very time consuming and error prone process.

*C. Using deep learning*

To overcome the disadvantages of above mentioned conventional methodologies new implementations based on deep learning algorithms replaced the previous methods[13] in recent years with increase in computing power and availability of standardized data sets and access to huge amount of data. Convolutional neural networks are the state of the art algorithms achieving highest accuracy rate.LENET architecture[14] was the first CNN architecture for traffic sign classification.

convolutional neural networks are biologically inspired multi-stage neural network architecture that learns the invariant features automatically. Each stage consists of filter bank(convolution) layer,non-linear transform layer,spatial pooling layer[15].The spatial pooling layer deceases the spatial information and acts like a complex cells in visual cortex.A gradient descent based optimizer is used for training and updating each filter to minimize loss function.

The output of all the layers is fed to the classifier for improving the accuracy of classification.

### III. DATA SET

German Traffic Sign Recognition Benchmark (GTSRB) dataset was described and visualized.It is a publicliy available dataset and it is created from 10 hours video of driving on different roads in Germany.video is captured using Prosilica GC 1380CH camera with framerate of 25 fps and the traffic sign extraction is performed using the NISYS Advanced Development and Analysis Framework(ADAF)[16] module based software system.

After cleaning and removing the redundant and repetitive frames the dataset is reduced to total 51,840 images of the 43 classes. All the images in the dataset have 32*32 size and the total dataset is divided into training data and testing data.Total 39,209 images are present as training data and 12,630 images as test data.

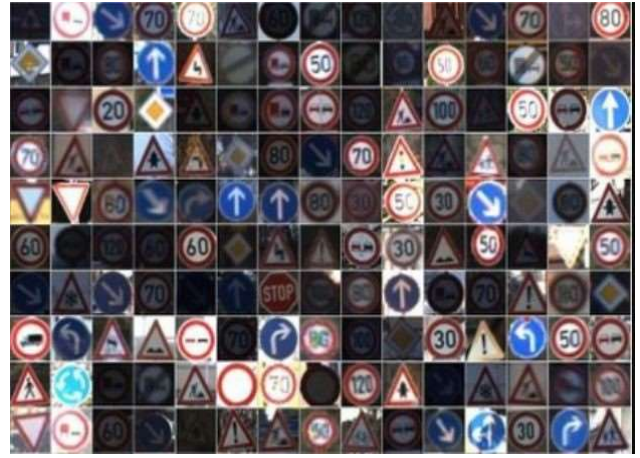

Fig. 1. Sample images from GTSRB dataset

### IV. CAPSULE NET ARCHITECTURE

Capsule networks consists of capsules rather than neurons. Capsule[17] is a group artificial neural networks that perform complicated internal computations on their inputs and encapsulate the results in a small vector. Each capsule captures the relative position of the object and if the object pose is changed then the output vector orientation is changed[18] accordingly making them equi-variant.

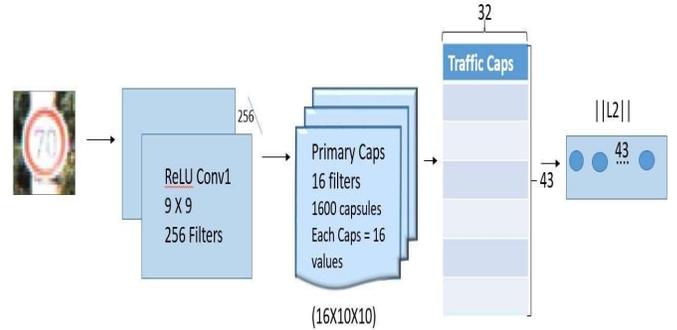

Fig. 2. Capsule sign architecture for traffic sign detection

Caps Net consists of multiple layers and the first layer is called primary capsules where each capsule receives a small part of the receptive field as input and tries to detect the pose of particular pattern. The output of the capsule is a vector and dynamic routing technique was used to ensure that the output is sent to the appropriate parent in the layer which can be inferred from fig. 2.

*A. Computation of capsule vector inputs and outputs*

The capsule computes a prediction vector[19] by multiplying the weight matrix($W_{ij}$) with its own output vector($u_i$). The coupling coefficient of that capsule corresponding output increases the scalar product and prediction[20] for that particular capsule output.

$$\hat{u}_{j|i} = W_{ij}u_i$$

where $u_{j|i}$=prediction vector, $W_{ij}$=weight matrix and $u_i$=output vector.

### B. Squash Function

In capsule networks a non-linear activation function called squashing function[21] is used. This function converts the length of the output vector into the probability of the capsule connecting to that object. It performs shrinking of the the long output vectors slightly below length one and short output vectors almost close to zero.

$$v_j = \frac{||s_j||^2}{1+||s_j||^2} \frac{s_j}{||s_j||}$$

where $s_j$=Total Input, $v_j$=Vector Output of capsule j.

### C. Routing Algorithm

Scalar output feature detectors of the CNN were replaced with vector output capsules[18] and max pooling with route by agreement. The dimensionality of the capsule increases with the increase In hierarchy because of the shift from place coding (encoded in continuous space))to rate coding(encoded in ) and the high level capsules represent entities which are complex and have more degree of freedom.This route by agreement is efficient than the max pooling used in the CNNs.

$$S_j = \sum_i c_{ij} \hat{u}_{j|i}$$

where $S_j$=summation matrix, $u_{j|i}$=prediction vector, and $c_{ij}$=coupling coefficients determined by iterative dynamic routing.

## V. EXPERIMENTS

### A. Data Preparation

The data is from the disk in pickled format.Pickling is the process of saving a file in a serialised format before writing it into the disk.Size of each image is 32*32 and total 34799 images are in the training dataset and 12630 in the testing dataset.

### B. preprocessing

The image brightness is enhanced with random uniform distribution of 0.6 to 1.5 and image contrast is also enhanced with random uniform distribution of 0.6 to 1.5.

The training dataset is augmented five-fold by replicating the available with data rotation ± 20, shear range of 0.2, width shift range of 0.2, horizontal flip which increased the training dataset size leading to better performance and regularizing thus avoiding the over fitting problem.

Augmenting the existing training dataset will increase the dataset size to 34799 X 5 = 173995 images.

### C. Network architecture

The architecture used for the traffic sign detection consists of the input layer and initially convolutional layers as part of primary capsules and the output vector of primary capsule is sent to traffic sign capsules.

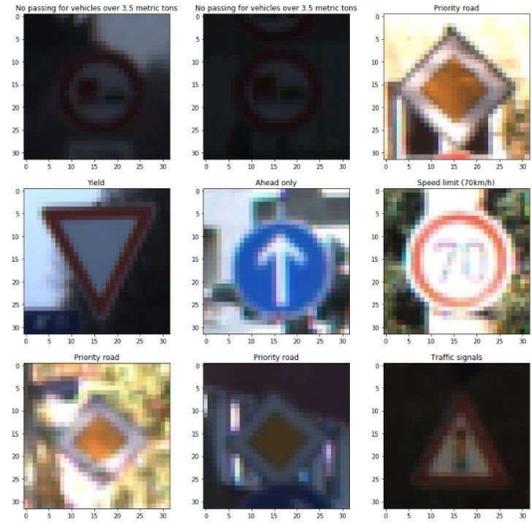

Fig. 3.    Visualizing the data

*1) Input Layer:* The input layer consists of input training images and the dimension is equal to the total training images.

*2) Primary Capsule Layer:* The first layer that follows the input layer is the primary capsule layer and for calculating the output first two convolution layers were used. The first convolution layer consists of kernel size 9 and 256 filters and padding was not used.Rectified Linear unit (ReLU) was used as the non linear activation function and a drop out of 0.7 which is fixed to be optimal after testing with different values.

Output is reshaped to get the output vectors of primary capsules.Since the primary capsule layer is fully connected to the traffic sign capsule layer the output vectors have to be squashed using the squashing function.Small epsilon value is added to the squash function to avoid the vanishing gradient problem while training.Now the output of the squash function is fed to the traffic sign capsule layer.

*3) Traffic Sign Capsule Layer:* To compute the output of traffic sign capsules,calculate the predicted output vectors for each and every primary,traffic sign capsule pair and implement the route by agreement algorithm.

The traffic sign capsule layer consists of 43 capsules each representing a particular class of the German traffic sign dataset with size of 32 each. For each capsule i in the first layer predict the corresponding weights and output vectors of every capsule j in the second layer.

### D. Reconstruction

A decoder network is added to the traffic sign capsule network which consists of fully connected network layer which helps in the reconstruction of the input images by tuning the output of the traffic sign capsule network.This feedback mechanism will make the network to preserve the information required for the reconstruction of the traffic sign across the entire network.This acts as regularization and this avoids the over-fitting of the data and helps in proper generalization of traffic signs.

*1) Mask:* For the reconstruction of the input traffic sign only that particular output vector corresponding to predicted traffic sign is sent and all remaining outputs should be masked.Masking function is used to avoid all the other output vectors during training phase.

The reconstruction mask is realized using the one-hot function.For the target class its value will be one and for all the other classes its value will be zero.

*2) Decoder:* The Decoder consists of a non linear activation layer of ReLU followed by a sigmoid activation layer.

### E. Losses

*1) Margin Loss:* The length of the instantiation output vector represent the probability of the respective capsule's entity exists or not.The digit class k has the longest vector output only if that traffic sign is present in the input image.

For every traffic sign capsule k the margin loss is separate and it is given as

$$L_k = T_k \max(0, m^+ - ||v_k||)^2 + \lambda(1-T_k)\max(0, ||v_k|| - m^-)^2$$

the value of $T_k$ is 1 if a traffic sign of class k is present and here m+ = 0.9 and m- = 0.1.$\lambda$ is a regularization parameter which stops the learning from shrinking the activity vector of all traffic sign capsules.

*2) Reconstruction Loss:* It is the difference between the squares of the input image and reconstructed image

$$R = (Input\ image)^2 - (Reconstructed\ image)^2$$

where R = Reconstruction Loss

*3) Final Loss:* The Final Loss is the sum of Margin loss and Reconstruction Loss scaled to a factor $\lambda$ which acts as a scaling factor and it should be very much less than one.

$$F = (Margin\ Loss) - \lambda(Reconstruction\ Loss)$$

where F= Final Loss, $\lambda$= 0.0005

Margin loss should always dominate the Reconstruction loss in comparison.If reconstruction loss is more in the final loss then the model tries to exactly match output image with the input image of training dataset which lead to overfitting of the model to the training data.

### F. Results

The model is evaluated using the testing data set of 12,630 testing images.Accuracy is computed as the ratio of the correctly identified traffic signs by the total number of traffic signs.[22]

$$Accuracy = \frac{\sum correctly\ identified\ traffic\ signs}{Total\ number\ of\ traffic\ signs}$$

with a batch size of 50 obtained an accuracy of 97.6 percent and a final loss of 0.0311028 evaluated on the testing dataset.

The performance evaluation is based on the correct classification rate(CCR) and binary loss ( 0 or 1 ) which means by counting the number of misclassification's.The test set is

TABLE I

COMPARISON OF CORRECT CLASSIFICATION RATE FOR DIFFERENT METHODS

| CCR (%) | Method |
|---|---|
| 97.62 | using Capsule networks |
| 96.14 | Random Forests [1] |
| 95.68 | LDA(HOG 2) [23] |
| 93.18 | LDA(HOG 1) [8] |
| 92.34 | LDA(HOG 3) [23] |

unbalanced in terms of the number of samples for a particular class but assume that all the classes are equally important and carry equal weightage.

Keras and tensor flow deep learning libraries with CUDA and CUDNN libraries for GPU accelerated training to implement this capsule network model. The model is trained using the system with Intel i7 7500U Processor 2.7G,3M Processor speed 8GB Memory RAM, 1 TB Hard Disk, NV GT 940 MX 2G DDR3 Nvidia GeForce GPU. It took 10 hours for training the model and can be decreased by using a high performance GPU like Nvidia K80.

From Table 1 we can conclude that capsule network is superior and performing better than other methods mentioned.

### G. Conclusion

Traffic sign detection is a challenging task and capsule networks using their inherent ability to detect the pose and spatial variances perform better when compared to CNN's and capsule networks increase the reliability and accuracy by correctly performing image classification and recognition tasks even on blurred,rotated and distorted images.

### H. Acknowledgement

Thanks to Dr. K.P.Soman for his constant motivation and for cultivating curiosity in deep learning.